\documentclass{ecai2006}
\usepackage{times}
\usepackage{graphicx}
\usepackage{latexsym}

\newcommand{\myproof}{\noindent {\bf Proof:\ \ }}
\newcommand{\myqed}{\mbox{$\diamond$}}
\newcommand{\myldots}{\mbox{$.  .$}}

\newcommand{\PRECEDENCE}{\mbox{\sc Precedence}}

\begin{document}

\title{Symmetry Breaking using Value Precedence}

\author{Toby Walsh\institute{National ICT Australia and University of
New South Wales, email: tw@cse.unsw.edu.au}}

\maketitle

\begin{abstract}
We present a comprehensive study of 
the use of value precedence constraints to break
value symmetry. 
We first give a simple encoding of value precedence into ternary
constraints that is both efficient and effective
at breaking symmetry. We then extend value precedence to deal with a 
number of generalizations like wreath value
and partial interchangeability. We also
show that value precedence is closely
related to lexicographical ordering. Finally, we 
consider the interaction between value
precedence and symmetry breaking
constraints for variable symmetries. 
\end{abstract}

\section{INTRODUCTION}

Symmetry is an important aspect
of many search problems. Symmetry
occurs naturally in many problems
(e.g. if we have two identical machines to schedule,
or two identical jobs to process).
Symmetry can also be introduced when we model a problem
(e.g. if we name the elements in a set, we
introduce the possibility of permuting their order). 
We must deal with symmetry or
we will waste much time visiting
symmetric solutions, as well as
parts of the search tree which
are symmetric to already visited 
parts. One simple but highly
effective mechanism to deal with symmetry is to
add constraints which eliminate symmetric solutions \cite{clgrkr96}.

Two common types of symmetry are variable
symmetries (which act just on variables),
and value symmetries (which act just on values).
With variable symmetries, 
we have a number of well understood symmetry
breaking constraints. For example, many
problems can be modelled using matrices
of decision variables, in which the rows
and columns of the matrix are symmetric and 
can be freely permuted. We can break 
such symmetry by lexicographically ordering 
the rows and columns  \cite{ffhkmpwcp2002}.
Efficient propagators have therefore
been developed for such ordering constraints \cite{fhkmwcp2002,lexchain}. 

Value symmetries are also common. However, symmetry
breaking for value symmetry is less well understood. 
In this paper, we study a common type of value 
symmetry where the values for variables are interchangeable. 
For example, if we assign orders to machines and two orders
are identical, we can swap them in any schedule. 
We show here how to deal with such symmetry. In particular, 
we give a simple encoding that breaks all
the symmetry introduced by interchangeable
values. We also show how this is closely
related to the lexicographical ordering constraint. 

\section{BACKGROUND}

A constraint satisfaction problem consists of a set of variables,
each with a domain of values, and a set of constraints
specifying allowed combinations of values for given subsets of
variables. A solution is an assignment of values to variables
satisfying the constraints.
Finite domain variables take values which are
integers, or tuples of integers taken from some given
finite set.
Set variables takes values which are sets of integers. 
A set variable $S$ has
a lower bound $lb(S)$ for its definite elements 
and an upper bound $ub(S)$ for its definite and
potential elements.

Constraint solvers typically explore partial assignments enforcing
a local consistency property. We consider 
the two most common local consistencies: arc consistency and bound consistency. 
Given a constraint $C$, 
a \emph{bound support} on $C$ is an
assignment of a value to each finite domain variable between its
minimum and  maximum and of a set 
to each set variable between its lower and upper bounds which
satisfies $C$. 
A constraint $C$ is \emph{bound consistent} (\emph{BC}) iff for each finite domain
variable, 
its minimum and maximum values belong to a bound support,
and for each set variable $S$, 
the values in $ub(S)$  belong to $S$ in at least one bound
support 
and the values in $lb(S)$ 
belong to $S$ in all bound supports. 
Given a constraint $C$ on finite domain variables,
a \emph{support} is assignment to each variable of a value in
its domain which satisfies $C$. 
A constraint $C$ on finite domains variables is \emph{generalized
  arc  consistent} (\emph{GAC})
iff for each variable, every value in its domain belongs to a support.

A \emph{variable symmetry} is a bijection on variables
that preserves solutions. For example, 
in a model of the rehearsal problem (prob039 in CSPLib)
in which we assign musical pieces to time slots, we can
always invert any schedule.
This is a reflection symmetry on the variables. 
A \emph{value symmetry} is a bijection on values
that preserves solutions. 
For example, in our model of the rehearsal problem,
two pieces requiring the same musicians
are interchangeable and can be freely permuted in any
solution. Note that some authors
call these {\em global} value
symmetries as they act globally on
values \cite{getree}. Finally, 
a pair of values are \emph{interchageable}
if we can swap them in any solution. 

\section{VALUE PRECEDENCE}

We can break all symmetry between a pair of interchangeable values
using a global value precedence constraint \cite{llcp2004}.
$$\PRECEDENCE([v_j,v_k],[X_1,\myldots ,X_n])$$ 
This holds iff $\min \{ i \ | \ X_i=v_j \vee i=n+1\} < 
 \min \{ i \ | \ X_i=v_{k} \vee i=n+2\}$. 
Law and Lee give a specialized
algorithm for enforcing GAC on such a global constraint \cite{llcp2004}. 
We show here that this is unnecessary. We can encode the constraint efficiently
and effectively into a simple sequence of ternary 
constraints. 

We introduce a sequence of 0/1 variables, 
where $B_i=1$ if $X_l=v_j$ for some $l<i$. 
Value precedence prevents us assigning $X_i=v_k$ unless $B_i=1$. 
To ensure this, we post the
sequence of ternary constraints, $C(X_i,B_i,B_{i+1})$
for $1 \leq i \leq n$ which hold iff
$X_i=v_j$ implies $B_{i+1}=1$, 
$X_i\neq v_j$ implies $B_i=B_{i+1}$,
and $B_i=0$ implies $X_i \neq v_k$. 
We also set $B_1=0$. 
We assume that we can enforce GAC on each individual 
ternary constraint $C$ using a table
constraint or using primitives like implication and equality
constraints. 
As the constraint graph is Berge-acyclic,
enforcing GAC on the ternary
constraints achieves GAC on 
$\PRECEDENCE([v_j,v_k],[X_1,\myldots ,X_n])$. 
Since $C$ is functional in its first two
arguments and there are $O(n)$ ternary
constraints, this takes $O(nd)$ time
where $d$ is the maximum domain
size of the $X_i$. This is therefore optimal
(as was Law and Lee's specialized algorithm). 
The incremental behavior is also good.
Law and Lee's algorithm 
maintains three pointers,
$\alpha$, $\beta$ and $\gamma$ to save re-traversing
the vector. 
A constraint engine will also 
perform well incrementally on this encoding
provided it ignores constraints once
they are entailed. Our experimental results 
support this claim. 

\section{MULTIPLE VALUE INTERCHANGEABILITY}

Many problems involve {\em multiple}
interchangeable values. For example, in a finite
domain model of the social golfer problem (prob010 in CSPLib)
in which we assign groups to golfers in each week, 
all values are interchangeable. 
To break all such symmetry, Law and Lee \cite{llcp2004}
propose the global constraint:
$$\PRECEDENCE([v_1,\myldots ,v_m],[X_1,\myldots ,X_n])$$
This holds iff 
$\min \{ i \ | \ X_i=v_i \vee i=n+1\} < 
 \min \{ i \ | \ X_i=v_{j} \vee i=n+2\}$ 
for all $1 \leq i < j < m$. 
To propagate this constraint, 
Law and Lee suggest decomposing it into pairwise precedence
constraints of the form 
$\PRECEDENCE([v_i,v_j],[X_1,\myldots ,X_n])$
for all $i<j$ \cite{llcp2004}. 
Law has conjectured (page 77 of \cite{lawthesis}),
that such a decomposition does not hinder
GAC propagation. We prove this is not the case. 

\begin{theorem}
Enforcing GAC on $\PRECEDENCE([v_1,\myldots ,v_m],$
$[X_1,\myldots ,X_n])$
is strictly stronger than enforcing GAC on 
$\PRECEDENCE([v_i,v_j],[X_1,\myldots ,X_n])$
for all $1 \leq i<j \leq m$.
\end{theorem}
\myproof
Clearly it is at least as strong. To show 
strictness, consider 
$\PRECEDENCE([1,2,3,4],[X_1,X_2,X_3,X_4])$
with $X_1 \in \{1\}$,
$X_2 \in \{1,2\}$,
$X_3 \in \{1,3\}$,
and $X_4 \in \{3,4\}$.
Then 
enforcing GAC on 
$\PRECEDENCE([1,2,3,4],[X_1,X_2,X_3,X_4])$
prunes 1 from the domain of $X_2$. 
However, $\PRECEDENCE([i,j],[X_1,X_2,X_3,X_4])$ is GAC
for all $1 \leq i<j \leq 4$.
\myqed

We propose instead a simple encoding
of $\PRECEDENCE([v_1,\myldots ,v_m],[X_1,\myldots ,X_n])$
into a sequence of ternary constraints.
We introduce $n+1$ finite domain variables, 
where $Y_i$ records the greatest index of the
values used so far in the precedence order. 
We then post a sequence of ternary constraints, 
$D(X_i,Y_i,Y_{i+1})$ 
for $1 \leq i \leq n$ which hold iff
$X_i \neq v_{j}$ for any $j > Y_i+1$,
    $Y_{i+1} = Y_i+1$ if $X_i = v_{1+Y_i}$ and
    $Y_{i+1} = Y_i$ otherwise.
We set $Y_1=0$. Again, we achieve GAC on the global
constraint simply by enforcing
GAC on the individual ternary constraints. 
This takes $O(nmd)$ time. By comparison, 
enforcing GAC on the decomposition into
precedence constraints between all pairs of 
interchangeable values takes $O(nm^2d)$ time. 

\section{PARTIAL INTERCHANGEABILITY}

We may have a partition on the values, 
and values within each partition are interchangeable. 
For example, in the model of the rehearsal problem
in which we assign musical pieces to time slots, we can
partition the musical pieces into those requiring the same musicians. 
Suppose the values are divided into $s$ equivalence
classes, then we can break all symmetry with 
the global constraint:
$$\PRECEDENCE([[v_{1,1},\myldots ,v_{1,m_1}],\myldots ,[v_{s,1},\myldots ,v_{s,m_s}]],[X_1,\myldots ,X_n])$$
This holds iff 
$\min \{ i \ | \ X_i=v_{j,k} \vee i=n+1\} < 
 \min \{ i \ | \ X_i=v_{j,k+1} \vee i=n+2\}$ 
for all $1 \leq j \leq s$ and $1 \leq k < m_j$. 
This global constraint can be decomposed into
$s$ precedence constraints, one for each equivalence class. 
However, such decomposition hinders propagation. 
\begin{theorem}
Enforcing GAC on 
$\PRECEDENCE([[v_{1,1},\myldots ,v_{1,m_1}],\myldots ,$
$[v_{s,1},\myldots ,v_{s,m_s}]],
[X_1,\myldots ,X_n])$
is strictly stronger than enforcing GAC on 
$\PRECEDENCE([v_{i,1},\myldots ,v_{i,m_i}],
[X_1,\myldots ,X_n])$
for $1\leq i \leq s$.
\end{theorem}
\myproof
Clearly it is at least as strong. To show 
strictness, consider 
$\PRECEDENCE([[1,2,3],[4,5,6]],[X_1,\myldots ,X_5])$
with $X_1, X_2, X_3 \in \{1,2,3,4,5,6\}$,
$X_4 \in \{3\}$ and $X_5 \in \{6\}$. 
Then 
$\PRECEDENCE([[1,2,3],[4,5,6]],[X_1,\myldots ,X_5])$
is unsatisfiable. 
However, $\PRECEDENCE([1,2,3],[X_1,\myldots ,X_5])$ and
$\PRECEDENCE([4,5,6],[X_1,\myldots ,X_5])$ are both GAC. 
\myqed

Decomposition is again unnecessary as we can encode 
the global constraint into a sequence of ternary constraints. 
The idea is to keep a tuple recording the greatest value
used so far within each equivalence class as we slide
down the sequence. We introduce $n+1$ new finite-domain variables, $Y_i$
whose values are $s$-tuples. We write $Y_i[j]$ to indicate
the $j$th component of the tuple. We then post  a sequence
of ternary constraints, $E(X_i,Y_i,Y_{i+1})$
for $1 \leq i \leq n$ which hold
iff for all $1 \leq j \leq s$ 
we have $X_i \neq v_{j,k}$ for all $k > Y_i[j]+1$,
$Y_{i+1}[j] = Y_i[j]+1$ if $X_i = v_{j,Y_i[j]+1}$ and
$Y_{i+1}[j] = Y_i[j]$ otherwise.
The value taken by $Y_{i+1}[j]$ is the largest index within
the $j$th equivalence class used up to $X_i$. 
Since $E$ is functional in its third argument, this takes $O(nde)$ 
time where $e = \prod_{i \leq s} m_i$. 
Note that if all values are interchangeable with
each other, then $s=1$ and $m_1=m$, and 
enforcing GAC takes 
$O(nmd)$ time. 
Similarly, for just one pair 
of interchangeable values, $s=n-1$, $m_1=2$ and $m_i=1$ for $i>1$,
and enforcing GAC takes $O(nd)$ time. 

\section{WREATH VALUE INTERCHANGEABILITY}

Wreath value interchangeability \cite{sellmann2}
is a common type of symmetry in problems where variables are assigned
a pair of values from $D_1 \times D_2$,
values in $D_1$ are fully interchangeable,
and for a fixed value in $D_1$, values in $D_2$ are
interchangeable as well. 
For example, if we are scheduling
a conference, the days of the week might be interchangeable,
and given a particular day, the meeting rooms might
then be interchangeable.   
For simplicity, we assume the same precedence
ordering is used for the values in $D_2$ for
every fixed value in $D_1$. 
However, we can 
relax this assumption without difficulty. 

We can break all the symmetry of wreath-value interchangeability
with the global constraint:
$$\PRECEDENCE([u_{1},\myldots ,u_{m},[v_{1},\myldots ,v_{p}]],[X_1,\myldots ,X_n])$$
This holds iff 
$\min \{ i \ | \ X_i[1]= u_{j} \vee i=n+1\} < 
 \min \{ i \ | \ X_i[1]= u_{j+1} \vee i=n+2\}$ 
for all $1 \leq j < m$,
and
$\min \{ i \ | \ X_i= \langle u_j,v_k \rangle \vee i=n+1\} < 
 \min \{ i \ | \ X_i= \langle u_j,v_{k+1} \rangle \vee i=n+2\}$ 
for all $1 \leq j \leq m$
and $1 \leq k < p$.
This can be decomposed into
precedence constraints of
the form $\PRECEDENCE([\langle u_i,v_j \rangle, \langle u_k,v_l \rangle],
[X_1,\myldots ,X_n])$, but this hinders
propagation. 
\begin{theorem}
Enforcing GAC on 
$\PRECEDENCE([u_{1},\myldots ,u_{m},$
$[v_{1},\myldots ,v_{p}]],$
$[X_1,\myldots ,X_n])$
is strictly stronger than enforcing GAC on 
$\PRECEDENCE([\langle u_i,v_j \rangle, \langle u_k,v_l \rangle],
[X_1,\myldots ,X_n])$
for all $1 \leq i < k \leq m$, and
for all $i=j$, $1 \leq i, j \leq m$, 
$1 \leq k < l \leq p$.
\end{theorem}
\myproof
Clearly it is at least as strong. To show 
strictness, consider 
$\PRECEDENCE([1,2,[3,4]],[X_1,X_2,X_3,X_4])$
with $X_1 \in \{ \langle 1,3 \rangle \}$,
$X_2 \in \{ \langle 1,3 \rangle,  \langle 1,4 \rangle \}$,
$X_3 \in \{ \langle 1,3 \rangle,  \langle 2,3 \rangle \}$,
$X_4 \in \{ \langle 2,3 \rangle,  \langle 2,4 \rangle \}$,
Then enforcing GAC on 
$\PRECEDENCE([[1,2,[3,4]],[X_1,X_2,X_3,X_4])$
prunes $\langle 1,3 \rangle$ from $X_2$.
However, 
$\PRECEDENCE([\langle u,v \rangle, \langle w,z \rangle],
[X_1,X_2,X_3,X_4])$ is GAC
for all $1 \leq u < w \leq 2$ and $3 \leq v, z \leq 4$, 
and for all $u=w$, $1 \leq u, w \leq 2$, $3 \leq v < z \leq 4$.
\myqed

We again can propagate such a precedence constraint using
a simple encoding into ternary constraints. 
We have a finite domain variable which records the greatest pair
used so far down the sequence. We can then encode 
$\PRECEDENCE([u_{1},\myldots ,u_{m},[v_{1},\myldots ,v_{p}]],
[X_1,\myldots ,X_n])$
by means of a sequence of ternary constraints, 
$F(X_i,Y_i,Y_{i+1})$ for $1 \leq i \leq n$ which
hold iff 
$X_i[1] \neq u_{j}$ for all $j > Y_i[1]+1$,
if $X_i[1] = u_{Y_i[1]}$ then $X_i[2] \neq v_j$
for all $j > Y_i[2]+1$,
$Y_{i+1} = \langle Y_i[1]+1,1 \rangle$ 
if $X_i[1] = u_{Y_i[1]}+1$ and $Y_i[2]=m$,
$Y_{i+1} = \langle Y_i[1],Y_i[2]+1 \rangle$ 
if $X_i = \langle u_{Y_i[1]},v_{Y_i[2]}+1\rangle$,
and $Y_{i+1} = Y_i$ otherwise.
Enforcing GAC using this encoding takes $O(ndmp)$ time. 
The extension to wreath value partial interchangeability
and to wreath value interchangeability
over $k$-tuples where $k>2$ are both straight forwards. 

\section{MAPPING INTO VARIABLE SYMMETRY}

An alternative way to deal with value symmetry
is to convert it into variable symmetry  \cite{ffhkmpwcp2002,llsac2005}. 
We introduce a matrix of 0/1
variables where $B_{ij}=1$ 
iff $X_i=j$. 
We assume the columns of this
 0/1 matrix represent
the interchangeable values, and the rows represent the original
finite domain variables. We now prove
that value precedence is equivalent to
lexicographically ordering the columns of this 0/1 matrix,
but that channelling into 0/1 variables 
hinders propagation. 

\begin{theorem}
$\PRECEDENCE([v_{1},\myldots ,v_{m}],[X_1,\myldots ,X_n])$
is equivalent to 
$X_i=v_j$ iff $B_{i,j}=1$ for $0 < j \leq m$
and $0 < i \leq n$, and
$[B_{1,j},\myldots ,B_{n,j}] \geq_{\rm lex} 
 [B_{1,j+1},\myldots ,B_{n,j+1}]$ 
for $0 < j < m$.
\end{theorem}
\myproof
By induction on $n$. 
In the base case, $n=1$,
$X_1=v_1$, $B_{1,1}=1$ and
$B_{1,j}=0$ for $1 < j \leq m$.
In the step case, suppose 
the value precedence constraint
holds for a ground assignment in which
$v_k$ is the largest value used so far. 
Consider any extension with $X_{n+1}=v_{l}$.
There are two cases. In the first,
$l=k+1$. The $l$th column is thus $[0,\myldots ,0,1]$. 
This is lexicographically less
than all previous columns. 
In the other case, 
$l\leq k$. Adding a new row with a single 
1 in the $l$th column
does not change the ordering between the 
$l-1$th, $l$th and $l+1$th columns. 
The proof reverses in a similar way.
\myqed

We could thus impose value precedence by channelling
into an 0/1 matrix model and using 
lexicographical ordering constraints \cite{ffhkmpwcp2002}.
However, this decomposition hinders propagation
as the lexicographical ordering constraints do not exploit
the fact that the 0/1 matrix has a single non-zero entry per row. 
Indeed, even if we add this implied constraint to the decomposition,
propagation is hindered. It is thus worth developing
a specialized propagator for the global \PRECEDENCE\
constraint. 

\begin{theorem}
GAC on $\PRECEDENCE([v_{1},\myldots ,v_{m}],[X_1,\myldots ,X_n])$
is strictly stronger than GAC on 
$X_i=v_j$ iff $B_{i,j}=1$ for $0 < j \leq m$ and $0 < i \leq n$, 
GAC on $[B_{1,j},\myldots ,B_{n,j}] \geq_{\rm lex} 
 [B_{1,j+1},\myldots ,B_{n,j+1}]$ 
for $0 < j < m$, and GAC on
$\sum_{j=1}^m B_{i,j}=1$ for $0 < i \leq n$.
\end{theorem}
\myproof
Clearly it is as strong. To show
strictness, consider $X_1=1$, $X_2 \in \{1,2,3\}$, $X_3=3$,
$B_{1,1}=B_{3,3}=1$, 
$B_{1,2}=B_{1,3}=B_{3,1}=B_{3,2}=0$, 
and $B_{2,1}, B_{2,2}, B_{2,3} \in \{0,1\}$. 
Then the decomposition is GAC. However,
enforcing GAC on the value precedence constraint
will prune 1 and 3 from $X_2$. 
\myqed

It is not hard to show that partial value interchangeability
corresponds to partial column
symmetry in the corresponding 0/1 matrix model.
As with full interchangeability, we
obtain more pruning with a specialized propagator
than with lexicographical ordering constraints. 

\section{SURJECTION PROBLEMS}

A surjection problem is one in which
each value is used at least once. Puget 
converts value symmetries
on surjection problems into variable
symmetries by channelling into
dual variables which record the
first index using a value \cite{pcp05}. 
For interchangeable values, this gives
$O(nm)$ binary symmetry breaking constraints:
$X_i=j \rightarrow Z_j \leq i$,
$Z_j=i \rightarrow X_i =j$,
and $Z_k < Z_{k+1}$ for all $1 \leq i \leq n$,
$1 \leq j \leq m$ and $1 \leq k < m$. 
Any problem can be made into a surjection
by introducing $m$ additional new variables
to ensure each value is used once. 
In this case, Puget's symmetry breaking 
constraints ensure value precedence. However,
they may not prune all possible values. Consider 
$X_1=1$, 
$X_2 \in \{1,2\}$,
$X_3 \in \{1,3\}$,
$X_4 \in \{3,4\}$,
$X_5=2$, $X_6=3$, $X_7=4$,
$Z_1=1$, 
$Z_2 \in \{2,5\}$,
$Z_3 \in \{3,4,6\}$, and
$Z_4 \in \{4,7\}$. 
Then all the binary implications
are AC. However, enforcing 
GAC on $\PRECEDENCE([1,2,3,4],[X_1,\ldots,X_7])$
will prune 1 from $X_2$. 

\section{SET VARIABLES}

We also meet interchangeable values
in problems involving set variables.
For example, in a set variable model of the social
golfers problem in which we assign a set of golfers
to the groups in each week, all values are interchangeable.
We can break all such symmetry with value
precedence constraints. 
For set variables, $\PRECEDENCE([v_1,\myldots ,v_m],[S_1,\myldots ,S_n])$
holds iff 
$\min \{ i \ | \ (v_j \in S_i \wedge v_k \not\in S_i) \vee i=n+1 \} < 
 \min \{ i \ | \ (v_k \in S_i \wedge v_j \not\in S_i) \vee i=n+2 \}$ 
for all $1 \leq j < k \leq m$ \cite{llcp2004}. 
That is, the first time 
we distinguish between $v_j$ and $v_k$
(because both values don't occur in a given
set variable), we have $v_j$ occurring and not $v_k$. 
This breaks all symmetry as we cannot now swap $v_j$ for $v_k$. 
Law and Lee again give a specialized propagator for
enforcing BC on $\PRECEDENCE([v_j,v_k],[S_1,\myldots ,S_n])$.
We prove here that this decomposition hinders propagation. 

\begin{theorem}
Enforcing BC on $\PRECEDENCE([v_1,\myldots ,v_m],[S_1,\myldots ,S_n])$
is strictly stronger than enforcing BC on 
$\PRECEDENCE([v_i,v_j],[S_1,\myldots ,S_n])$
for all $1 \leq i<j \leq m$.
\end{theorem}
\myproof
Clearly it is at least as strong. To show 
strictness, consider 
$\PRECEDENCE([0,1,2],[S_1,S_2,S_3,S_4,S_5])$
with $\{\} \subseteq S_1 \subseteq \{0\}$,
$\{\} \subseteq S_2 \subseteq \{1\}$,
$\{\} \subseteq S_3 \subseteq \{1\}$,
$\{\} \subseteq S_4 \subseteq \{0\}$,
and
$S_5 = \{2\}$.
Then enforcing BC on 
$\PRECEDENCE([0,1,2],[S_1,S_2,S_3,S_4,S_5])$
sets $S_1$ to $\{0\}$. 
However, $\PRECEDENCE([i,j],[S_1,S_2,S_3,S_4,S_5])$ is BC
for all $0 \leq i<j \leq 2$.
\myqed

As with finite domain variables, we do 
need to introduce a new propagator 
nor to decompose this global constraint. 
We view each set variable in terms of 
its characteristic function (a vector of
0/1 variables). This gives us an $n$ by $d$ matrix
of 0/1 variables with column symmetry in the $d$
dimension. Unlike the case with finite domain variables,
rows can now have any sum. We can break all such column symmetry
with a simple lexicographical ordering constraint \cite{ffhkmpwcp2002}.
If we use the lex chain propagator \cite{lexchain},
we achieve BC on the original value
precedence constraint in $O(nd)$ time. 

In many constraint solvers, set variables also have restrictions
on their cardinality. Unfortunately, 
adding such cardinality information 
makes value precedence intractable to propagate. 
\begin{theorem}
Enforcing BC on 
$\PRECEDENCE([v_1,\myldots ,v_m],[S_1,\myldots ,S_n])$
where set variables have cardinality bounds is 
NP-hard. 
\end{theorem}
\myproof
We give a reduction from a 1-in-3 SAT problem in $N$ Boolean 
variables, $x_1$ to $x_N$ and $M$ positive clauses. 
We let $n=2N+M$, $m=2N$ and $v_i=i$. 
To encode the truth assignment which satisfies
the 1-in-3 SAT problem, we have
$S_{2i}= \{2i,2i+1\}$ and
$\{2i\} \subseteq S_{2i+1} \subseteq \{2i,2i+1\}$
for $1 \leq i \leq N$.
$S_{2i+1}$ will be $\{2i\}$ iff $x_i$ is false
and $\{2i,2i+1\}$ otherwise. 
The remaining $M$ CSP variables represent 
the $M$ clauses. 
Suppose the $i$th clause is $x_j \vee x_k \vee x_l$, 
We let $S_{2N+i} \subseteq \{2j+1, 2k+1, 2l+1\}$. 
Finally, we force $S_{2N+i}$ to take two of the
values $2j+1$, $2k+1$, $2l+1$ from its upper bound.
Value precedence only permits this if
exactly two out of $S_{2j}$, $S_{2k}$ and $S_{2l}$ take
the set value representing ``false''. 
The global value precedence constraint 
thus has bound
support iff the corresponding 1-in-3 SAT
problem is satisfiable. Hence, enforcing BC is NP-hard. 
\myqed

\section{VALUE AND VARIABLE SYMMETRY}

In many situations, we have both variable
and value symmetry. 
Can we safely combine
together symmetry breaking constraints
for variable symmetries and value symmetries? Do
we break all symmetry?

\subsection*{INTERCHANGEABLE VARIABLES}

\newcommand{\MONO}{\mbox{\sc IncreasingSeq}}
Suppose that all $n$ variables and $m$ values
are interchangeable.
We can safely combine a global value precedence constraint 
(which breaks all the value symmetry)
with a simple ordering constraint (which breaks all the
variable symmetry). However, this does not break
all symmetry. For example, $[1,2,2]$ and 
$[1,1,2]$ are symmetric since 
inverting the first sequence and permuting 1
with 2 gives the second sequence. However, both
sequences satisfy
the value precedence and ordering constraints. 
We can break all symmetry with the
global constraint $\MONO([X_1,\myldots ,X_n])$ which 
holds iff $X_1=v_1$, 
$X_{i+1} = X_{i}$ or ($X_{i} = v_j$ and $X_{i+1}=v_{j + 1}$)
for all $0 < i < n$, 
and $|\{ i \ | \ X_i=v_j \}| \leq  |\{ i \ | \ X_i=v_{j+1} \}|$
for all $0 < j < m$. 
That is, the values and the
number of occurrences of each value increase monotonically. 
We can also have the values increasing but 
the number of occurrences decreasing. 
One way to propagate this constraint
is to consider the corresponding 
0/1 matrix model. The $\MONO$ constraint
lexicographically orders the rows and columns, as
well as ordering the columns by their sums. 

Consider, now, (partially) interchangeable set variables
taking (partially) interchangeable values. 
This corresponds to an 0/1 matrix with
(partial) row and (partial) column symmetry. 
Unfortunately,  enforcing 
breaking all row and column symmetry is NP-hard \cite{bhhwaaai2004}.
We cannot expect therefore to break all symmetry 
when we have interchangeable set variables and 
interchangeable values. We can break symmetry
partially by lexicographical ordering the rows and columns of the
corresponding 0/1 matrix. 

\subsection*{MATRIX SYMMETRY}

Variables may be arranged in a matrix which
has row and/or column symmetry \cite{ffhkmpwcp2002}.
Lexicographical ordering constraints will break such symmetries. 
Suppose that values are also (partially) interchangeable. 
As lexicographical ordering constraints can be combined
in any number of dimensions \cite{ffhkmpwcp2002}, and
as value precedence is equivalent
to lexicographically ordering the 0/1 model,
we can safely combine 
value precedence and
row and column symmetry breaking constraints. 

\subsection*{VARIABLE REFLECTION SYMMETRY}

Suppose we have a sequence of $2n$ variables with
a reflection symmetry. 
Then we can break all such symmetry with the 
lexicographical
ordering constraint: $[X_1,\ldots,X_n] \leq_{\rm lex} [X_{2n},\ldots,X_{n+1}]$. 
For an odd length sequence, we just miss out the
middle element. If values are also (partially) interchangeable, 
we can combine such a reflection symmetry breaking
constraint with precedence constraints. 
Whilst these symmetry breaking constraints are 
compatible, 
they do not break all symmetry. 
For example, $[1,2,1,1,2]$ and
$[1,2,2,1,2]$ are symmetric since
inverting the first sequence and permuting 1 with 2
gives the second sequence. 
However, both sequences satisfy all 
symmetry breaking
constraints. 

\subsection*{VARIABLE ROTATION SYMMETRY}

Suppose we have a sequence of $n$ variables with
a rotation symmetry. That is, if we 
rotate the sequence, we obtain a symmetric solution. 
We can break all such symmetry with the 
constraints: 
$[X_1,\ldots,X_n] \leq_{\rm lex} [X_{i},\ldots,X_{n},X_1, \ldots X_{i-1}]$
for $1<i \leq n$. 
If values are also (partially) interchangeable, then
we can combine such symmetry breaking constraints
with precedence constraints. 
Whilst these symmetry breaking constraints are 
compatible, 
they do not break all symmetry. 
For example, $[1,1,2,1,2]$ and
$[1,2,1,2,2]$ are symmetric since
rotating the first sequence by 2 elements and permuting 1 with 2
gives the second sequence. 
However, both sequences satisfy all 
symmetry breaking constraints. 



\section{EXPERIMENTAL RESULTS}

To test the efficiency and effectiveness of these encodings
of value precedence constraints, we ran a range of experiments.
We report results here on Schur numbers (prob015 in CSPLib). This 
problem was used by Law and Lee in the first experimental
study of value precedence \cite{llcp2004}. We have observed similar
results in other domains like the social golfers problem
and Ramsey numbers (prob017 in CSPLib). 

The Schur number $S(k)$ is the largest integer 
$n$ for which the interval $[1,n]$ can be partitioned into $k$ 
sum-free sets. $S$ is sum-free iff $\forall a,b,c \in S \ . \
a \neq b+c$. Schur numbers are related to Ramsey numbers, $R(n)$ through
the identity: $S(n) \leq R(n)-2$. 
Schur numbers were proposed by the famous German mathematician Isaai Schur 
in 1918. $S(4)$ was open until 1961 when it 
was first calculated by computer. $S(3)$ is 13,
$S(4)$ is 44, and $160 \leq S(5) \leq 315$. 
We consider the corresponding decision problem,
$S(n,k)$ which asks if the interval $[1,n]$ can be partitioned into $k$ sum-free sets.
A simple model of this uses $n$ finite domain variables
with $k$ interchangeable values.

Results are given in Table 1. 
The model {\em all} uses a single global precedence
constraint to break all value symmetry.
The model {\em adjacent} uses the method proposed by Law and Lee in
\cite{llcp2004} which posts $O(k)$ precedence
constraints between adjacent interchangeable values.
The model {\em none} use no 
precedence
constraints. We coded the problem using the finite domain 
library in SICSTUS 3.12.3,
and ran it on an AMD Athlon Dual Core 2.2GHz processor
with 1 GB RAM.

\begin{table*}[htb]
\begin{center}
\begin{tabular}{|c|rrr|rrr|rrr|} \hline
problem  & \multicolumn{9}{c|}{value symmetry breaking} \\
$S(n,k)$ & \multicolumn{3}{c|}{none} & \multicolumn{3}{c|}{adjacent values} & \multicolumn{3}{c|}{all values} \\ 
  & {\bf c} & {\bf b} & {\bf t}   & {\bf c} & {\bf b} & {\bf t}   & {\bf c} & {\bf b} & {\bf t}  \\
\hline
$S(13,3)$ &  126 &  276 & 0.02 &      360 &   {\bf 46} &  {\bf 0.01} &  243 & {\bf  46} &  {\bf 0.01} \\
$S(13,4)$ &  126 & 134400 & 16.43 &   477 & {\bf 2,112} & 1.48 &  243 & {\bf 2,112} &  {\bf 0.85} \\
$S(13,5)$ & & &                   &   777 & 210,682 & 20.80 &  243 & {\bf 6,606} & {\bf 11.88} \\
$S(13,6)$ & & &                   &   879 & 309,917 & 79.60 &  243 & {\bf 1,032} & {\bf 42.51} \\
\hline
$S(14,3)$ &  147 & 456 &  0.02   &     399 & {\bf 76} &  {\bf 0.02} &   273 &  {\bf 76} &   {\bf 0.02} \\
$S(14,4)$ &  147 & 46,1376 & 39.66 &   525 & {\bf 8,299}  & 3.06 &  273 &  {\bf 8,299} &  {\bf 1.96} \\
$S(14,5)$ & & &                    &  816  & 813,552 & 66.83 & 273 & {\bf 58,558} & {\bf 40.35} \\
$S(14,6)$ & & &                    & 1,731 & 250,563 & 348.06 & 273 & {\bf 57,108} & {\bf 197.39} \\
\hline
$S(15,3)$ & 168 & 600 & 0.03 &        438 & {\bf 100} &  {\bf 0.02} &   303 &   {\bf 100} &  {\bf 0.02} \\
$S(15,4)$ & 168 & 1,044,984 & 101.36  & 573 & {\bf 17,913} & 7.92 &  303 & {\bf 17,913} &  {\bf 4.73} \\
$S(15,5)$ & & &                       & 855 & 1,047,710 & 259.15 & 303 & {\bf 194,209} & {\bf 145.97} \\
$S(15,6)$ & & & & & & & & & \\ \hline
\end{tabular}
\caption{Decision problem associated with Schur numbers:
{\bf c}onstraints posted, {\bf b}acktracks and
{\bf t}imes to find all solutions in seconds
to $S(n,k)$. Blank entries are for problems not solved
within the 10 minute cut off. Results are similar to find first
solution. } 
\end{center}
\end{table*}

The results show the benefits of a global
value precedence constraint. 
With a few interchangeable values, 
we see the same pruning 
using {\em adjacent} as {\em all}. 
However, we observe better runtimes with the {\em all} model
as it encodes into fewer ternary
constraints ($O(n)$ versus $O(nk)$). 
With more interchangeable values
(e.g. $k>4$), we observe both better
runtimes and more pruning with the
single global precedence constraint in the {\em all} model. 
The encoding of this global constraint into
ternary constraints appears therefore to be an efficient
and an effective mechanism to deal with 
interchangeable values. 

\section{RELATED WORK}

Whilst there has been much work on
symmetry breaking constraints for variable symmetries,
there has been less on value symmetries.
Law and Lee formally 
defined value precedence 
\cite{llcp2004}. 
They also gave specialized propagators for breaking value
precedence for a pair of interchangeable values. 
Gent proposed the first encoding of value precedence constraint 
\cite{gentvalsymm}. 
However, it is uncertain what consistency is achieved
as the encoding indexes with finite domain variables. 

A number of methods 
that modify the underlying solver have been proposed
to deal with value symmetry.
Van Hentenryck {\it et al.} gave a 
labelling schema for
breaking all symmetry with interchangeable values
\cite{hafpijcai2003}. 
Inspired by this method, 
Roney-Dougal {\it et al.} 
gave a polynomial method to construct 
a GE-tree, a search tree without value symmetry \cite{getree}. 
Finally, Sellmann and van Hentenryck gave
a $O(nd^{3.5}+n^2d^2)$
dominance detection algorithm
for breaking all symmetry when both 
variables and values are interchangeable
\cite{sellmann2}. 

There are a number of earlier (and related) results
about the tractability of symmetry breaking. 
Crawford {\it et al.} prove that breaking
all symmetry in propositional problems is NP-hard
in general \cite{clgrkr96}.
Bessiere {\it et al.} prove that
the special case of breaking all row and column symmetry for
variables in a matrix model is NP-hard \cite{bhhwaaai2004}.
Sellmann and van Hentenryck prove a closely
related result that dominance detection 
for breaking all symmetry with set
variables and values that are interchangeable
is NP-hard \cite{sellmann2}. 

\section{CONCLUSIONS}

We have presented a detailed study of 
the use of value precedence constraints to break
value symmetry. 
We first gave a simple encoding of value
precedence into ternary constraints that is both efficient and effective.
We then extended
value precedence to deal with a 
number of generalizations like wreath value
and partial interchangeability. We have
also shown how value precedence is closely
related to lexicographical ordering. Finally, we 
considered the interaction between value
precedence and other symmetry breaking
constraints. 
There are a number of interesting open questions.
For example, how does value precedence
interact with variable and value ordering heuristics? 


\bibliographystyle{ecai2006}


\end{document}